\documentclass[11pt]{article}
\usepackage{jair, rawfonts}

\usepackage[colorlinks,urlcolor=blue,linkcolor=blue,citecolor=blue]{hyperref}
\usepackage{stmaryrd,scalerel}
\usepackage{color,array}
\usepackage{graphicx}
\usepackage{scalerel}
\usepackage{tikz}
\usetikzlibrary{svg.path}
\usepackage{xcolor}
\usepackage{amsmath}

\usepackage{amsfonts}
\usepackage{mathtools}
\usepackage[numbers]{natbib}
\usepackage[ruled,vlined,noend,linesnumbered]{algorithm2e}
\DontPrintSemicolon
\usepackage{multirow}
\usepackage{multicol}
\usepackage{booktabs}
\usepackage{blkarray}
\usepackage{fancyhdr}
\usepackage{float}
\usepackage[verbose]{placeins} 
\usepackage{wrapfig}
\usepackage{subcaption}

\definecolor{orcidlogocol}{HTML}{A6CE39}
\tikzset{
    orcidlogo/.pic={
        \fill[orcidlogocol] svg{M256,128c0,70.7-57.3,128-128,128C57.3,256,0,198.7,0,128C0,57.3,57.3,0,128,0C198.7,0,256,57.3,256,128z};
        \fill[white] svg{M86.3,186.2H70.9V79.1h15.4v48.4V186.2z}
        svg{M108.9,79.1h41.6c39.6,0,57,28.3,57,53.6c0,27.5-21.5,53.6-56.8,53.6h-41.8V79.1z M124.3,172.4h24.5c34.9,0,42.9-26.5,42.9-39.7c0-21.5-13.7-39.7-43.7-39.7h-23.7V172.4z}
        svg{M88.7,56.8c0,5.5-4.5,10.1-10.1,10.1c-5.6,0-10.1-4.6-10.1-10.1c0-5.6,4.5-10.1,10.1-10.1C84.2,46.7,88.7,51.3,88.7,56.8z};
    }
}

\newcommand\orcidicon[1]{\href{https://orcid.org/#1}{\mbox{\scalerel*{
                \begin{tikzpicture}[yscale=-1,transform shape]
                \pic{orcidlogo};
                \end{tikzpicture}
            }{|}}}}

\jairheading{}{}{}{}{}
\ShortHeadings{COMFORT}
{Li \& Jha}

\begin{document}

\title{COMFORT: A Continual Fine-Tuning Framework for Foundation Models Targeted at Consumer Healthcare} 
\author{\name Chia-Hao Li $^{\textsuperscript{\orcidicon{0000-0001-9557-6050}}}$ \email chli@princeton.edu \\
        \addr Dept. of Electrical \& Computer Engineering, Princeton University\\
        Princeton, NJ 08544 USA\\
        \name Niraj K. Jha $^{\textsuperscript{\orcidicon{0000-0002-1539-0369}}}$, \textit{Fellow, IEEE} \email jha@princeton.edu \\
        \addr Dept. of Electrical \& Computer Engineering, Princeton University\\
        Princeton, NJ 08544 USA}
\maketitle

\begin{abstract}
Wearable medical sensors (WMSs) are revolutionizing smart healthcare by enabling continuous, real-time monitoring of 
user physiological signals, especially in the field of consumer healthcare. The integration of WMSs and modern machine 
learning (ML) enables unprecedented solutions to efficient early-stage disease detection. Despite the success of
Transformers in various fields, their application to sensitive domains, such as smart healthcare, remains underexplored 
due to limited data accessibility and privacy concerns. To bridge the gap between Transformer-based foundation models 
and WMS-based disease detection, we propose COMFORT, a continual fine-tuning framework for foundation models 
targeted at consumer healthcare. COMFORT introduces a novel approach for pre-training a Transformer-based foundation 
model on a large dataset of physiological signals exclusively collected from healthy individuals with commercially 
available WMSs. We adopt a masked data modeling (MDM) objective to pre-train this health foundation model.  We then 
fine-tune the model using various parameter-efficient fine-tuning (PEFT) methods, such as low-rank adaptation (LoRA) 
and its variants, to adapt it to various downstream disease detection tasks that rely on WMS data. In addition, 
COMFORT continually stores the low-rank decomposition matrices obtained from the PEFT algorithms to construct a 
library for multi-disease detection. The COMFORT library enables scalable and memory-efficient disease detection on 
edge devices. Our experimental results demonstrate that COMFORT achieves highly competitive performance while
reducing memory overhead by up to 52\% relative to conventional methods. Thus, COMFORT paves the way for personalized 
and proactive solutions to efficient and effective early-stage disease detection for consumer healthcare.
\footnote[2]{\scriptsize This work has been submitted to the ACM for possible publication. Copyright may be transferred without notice, after which this version may no longer be accessible.\hfill} \footnote[3]{\scriptsize \copyright~2024 ACM. Personal use of this material is permitted. Permission from ACM must be obtained for all other uses, in any current or future media, including reprinting/republishing this material for advertising or promotional purposes, creating new collective works, for resale or redistribution to servers or lists, or reuse of any copyrighted component of this work in other works.\hfill}
\end{abstract}

\section{Introduction}
Physical and mental illnesses not only negatively impact individual well-being but also adversely impact the global 
society. Therefore, the pursuit of efficient and effective early-stage disease detection has become a crucial 
research area in the consumer healthcare domain. Fortunately, modern advances in machine learning (ML) and wearable 
medical sensors (WMSs) highlight unprecedented ways to address this challenge. WMSs have revolutionized the healthcare 
field through passive, continuous, non-invasive, and real-time monitoring of physiological signals. These devices can 
collect a wealth of data, including galvanic skin response, heart rate, skin temperature, blood pressure, oxygen 
saturation, and other vital signs. A sophisticated ML model can then be trained on such data to enable efficient and 
accurate inference for early-stage disease detection. The integration of ML and WMSs promises to streamline the 
disease detection process, making it feasible even in out-of-clinic scenarios \cite{covidD, diabD, mhD, doctor, 
arrhythmia, page, parkinson, MedAI, SAS}. 

On the other hand, Transformers \cite{transformer} are revolutionizing various research fields, such as natural 
language processing (NLP) \cite{nlp_survey, sentiment, machine_tranlate, QA} and computer vision 
(CV) \cite{ViT_survey, ViT4CV, ViT}. The unique self-attention mechanism in Transformer models processes complex 
sequential data in parallel, which significantly reduces training time. As a result, numerous intensively pre-trained 
large language models (LLMs) \cite{gpt-3, gpt-4, bert, llama, llama-2} and large vision models (LVMs) 
\cite{ViT, clip, convnet} have been pre-trained as foundation models. These models are designed to perform 
general-purpose tasks and can be fine-tuned for specific downstream applications. They are pre-trained on an enormous 
amount of open-source unlabeled data through self-supervised learning to extract general knowledge from the target 
domain. However, fine-tuning large foundation models for every new task is often impractical and resource-intensive. 
Therefore, various parameter-efficient fine-tuning (PEFT) methods \cite{lora, dora, cola, adapter1, adapter2, prefix1} 
have been proposed to address this challenge. In general, these methods selectively tune a small subset of parameters 
to reduce the computation burden and time required for fine-tuning. Hence, they enable foundation models to efficiently 
adapt to diverse applications while maintaining scalability.

Despite the significant success of foundation models in various fields, their capabilities in sensitive domains like 
smart healthcare remain largely underexplored \cite{health-llm}. Patient health data, including electronic health 
records, medical images, and physiological signals, are governed by numerous laws and restrictions to ensure data 
security and patient privacy. Consequently, retrieving sufficient general data to pre-train a foundation model for 
healthcare applications is challenging. Several multimodal LLMs, fine-tuned for smart healthcare, have demonstrated 
promising performance on understanding domain knowledge \cite{hllm_survey, hllm_1, hllm_2, hllm_3, hllm_4, hllm_5, 
hllm_6}. However, their primary focus is on developing conversational artificial intelligence (AI) assistants for 
medical support rather than on early-stage disease detection in consumer health tasks. Consumer health disease 
detection relies heavily on sequential time-series physiological signals collected from WMSs, which differ significantly 
from static linguistic text data. The unique characteristics of WMS data, such as high dimensionality, nonlinear 
relationships, and continuous nature, require foundation models to comprehend both individual data points and their 
dynamic patterns over time \cite{health-llm}. Therefore, developing Transformer-based foundation models using 
physiological data from WMSs for disease detection tasks poses new challenges.

To address the aforementioned challenge, we propose a framework called COMFORT. It introduces a novel approach to 
developing foundation models tailored to WMS data that can be fine-tuned across various downstream disease detection 
tasks in the consumer healthcare domain. We pre-train a Transformer-based foundation model using a large WMS dataset 
collected exclusively from healthy individuals. We employ a masked data modeling (MDM) objective for pre-training, 
guiding the foundation model to grasp the essence and dynamic patterns of physiological data from WMSs. Subsequently, 
the foundation model can be efficiently fine-tuned using a PEFT method and disease-specific WMS datasets gathered from 
patients for downstream disease detection tasks. COMFORT adopts a PEFT algorithm based on low-rank adaptation (LoRA) 
\cite{lora}. For our foundation model with a pre-trained weight matrix $W_0 \in \mathbb{R}^{d \times k}$, LoRA 
constrains its update $W_0 + \Delta W$ by representing the latter with a low-rank decomposition $W_0 + \Delta W = 
W_0 + BA$, where $B \in \mathbb{R}^{d \times r}$, $A \in \mathbb{R}^{r \times k}$, and the rank $r \ll \text{min}(d, k)$ 
for $\{r, d, k\} \in \mathbb{N}$ \cite{lora}.

\begin{figure}[t]
    \centering
    \begin{subfigure}[b]{0.5\textwidth}
        \centering
        \includegraphics[width=\textwidth]{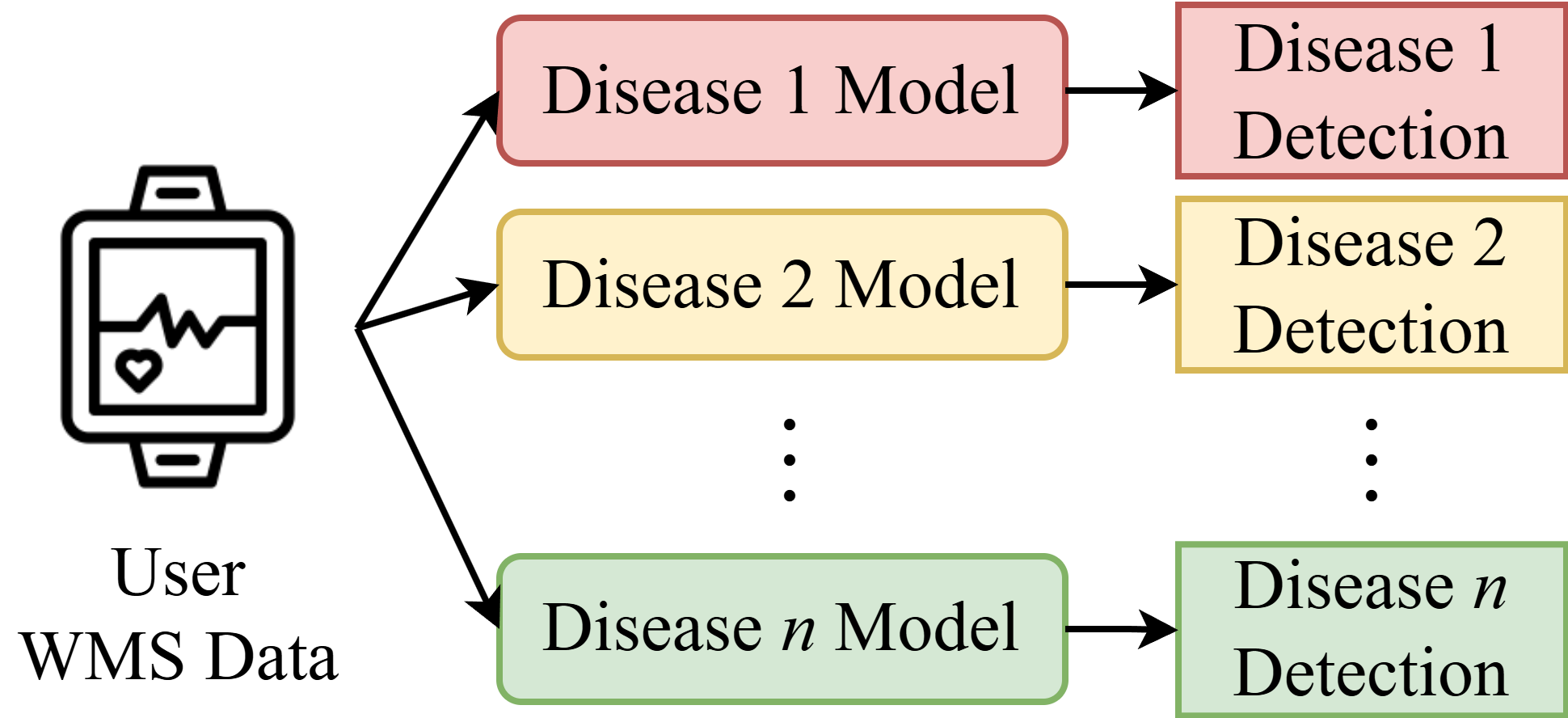}
        \caption{Conventional Disease Detection Methods}
    \end{subfigure}
    \begin{subfigure}[b]{0.7\textwidth}
        \centering
        \includegraphics[width=\textwidth]{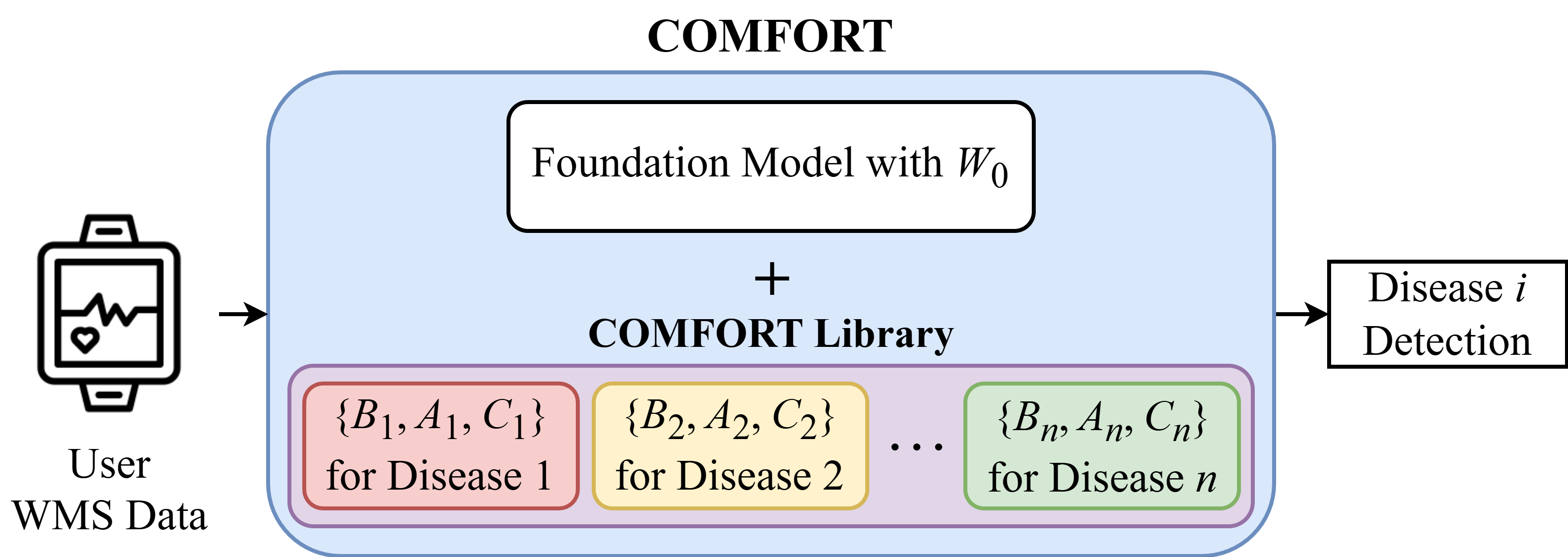}
        \caption{COMFORT Framework}
    \end{subfigure}
    \caption{An overview of frameworks for disease detection with WMS data in consumer health: (a) conventional disease detection methods, and (b) COMFORT framework. COMFORT flexibly leverages a library of weight variances to detect various diseases with a single foundation model, whereas conventional methods require a customized model for each disease. $W_0$ 
represents the pre-trained weights of the foundation model.}
    \label{fig:overview}
\end{figure}

COMFORT keeps the original pre-trained weights $W_0$ unchanged even after learning new tasks. It continually stores 
only the low-rank decomposition matrices 
obtained using the PEFT algorithm and the classifiers 
dedicated to learned tasks. This approach allows COMFORT to build a library of distinct low-rank weight matrices 
$\{B_i, A_i\}$ and classifiers $\{C_i\}$ for all learned diseases, where $i = \{1, 2, \ldots, n\}$. 
Using a single foundation model, end users can efficiently and flexibly employ different $\{B_i, A_i, C_i\}$ from the 
library to detect various diseases. As a result, compared to the conventional approach that employs separate models 
for distinct diseases, COMFORT significantly reduces memory and energy consumption on edge devices. 
Fig.~\ref{fig:overview} illustrates how COMFORT leverages its library to detect various diseases, contrasting it with 
conventional methods that require multiple models. To detect disease $i$, the user can download its corresponding 
$\{B_i, A_i, C_i\}$ from the library and apply it to the foundation model. The model then performs disease $i$ detection 
with model weights $W_0 + \Delta W_i = W_0 + B_iA_i$ and classifier $C_i$. Our results demonstrate COMFORT's potential 
for creating a scalable and adaptive framework for consumer health disease detection with WMS data, paving the way for 
personalized and proactive healthcare solutions.

The rest of the article is organized as follows. Section \ref{sec:background} discusses related works on 
Transformer-based foundation models, PEFT methods, and Transformer models employed in the healthcare domain. 
Section \ref{sec:COMFORT} gives details of our framework. Section \ref{sec:setup} provides information on the disease 
datasets used in our experiments and the implementation details. Section \ref{sec:experiments} presents experimental 
results. Section \ref{sec:discuss} discusses possible future work. Finally, Section \ref{sec:conclusion} concludes the 
article.

\section{Background and Related Work}
\label{sec:background}
In this section, we provide background material and discuss related works that will help readers understand the rest of 
the article. 

\subsection{Transformers and Foundation Models}

\begin{figure}[t]
    \centering
    \includegraphics[width=0.6\linewidth]{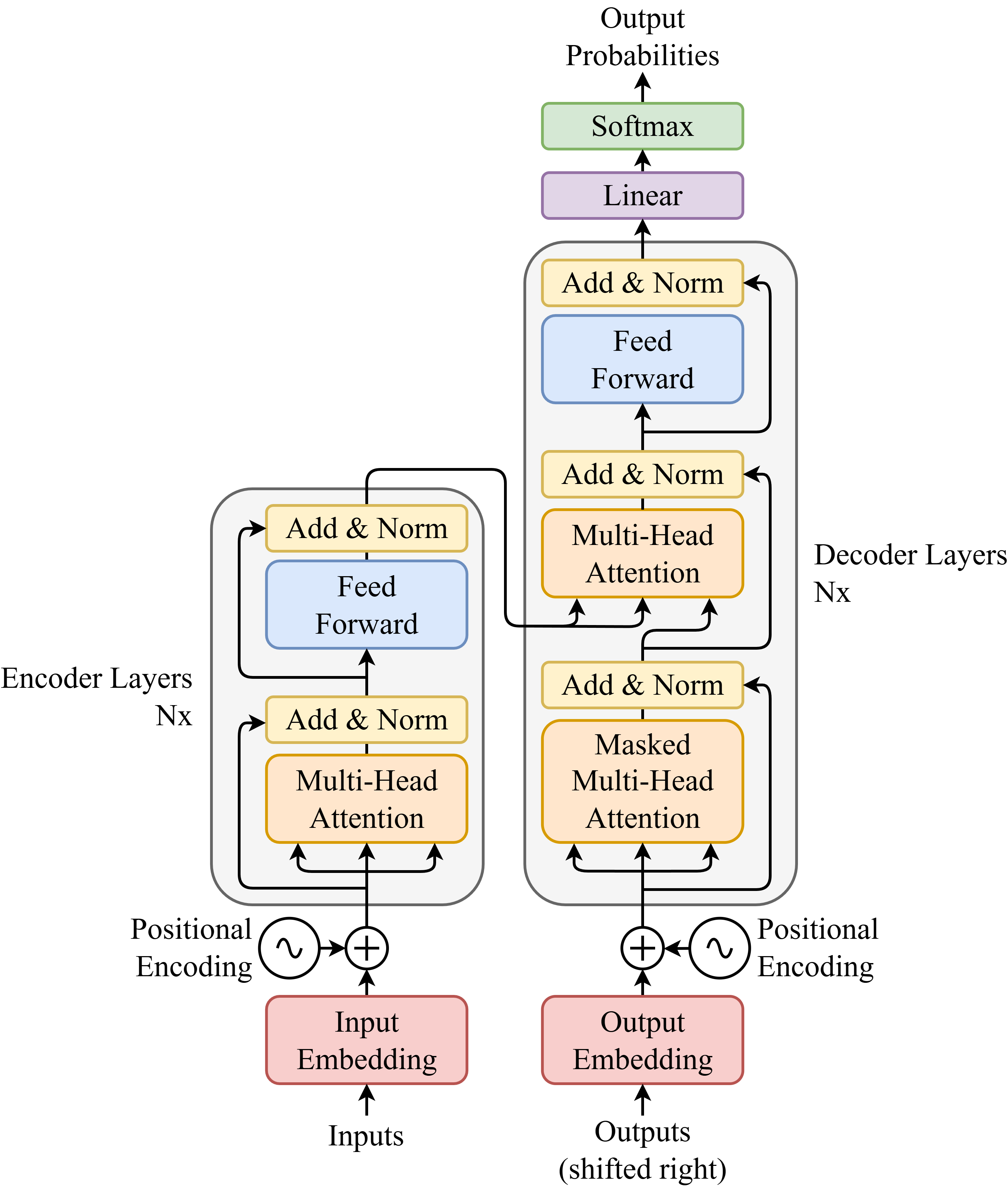}
    \caption{The Transformer architecture. This figure is remade from \cite{transformer}.}
    \label{fig:transformer}
\end{figure}

A Transformer is a model architecture that relies on the self-attention mechanism \cite{attention}. The original
Transformer model \cite{transformer} was designed for machine translation tasks. Fig.~\ref{fig:transformer} shows
the Transformer model architecture. It generally consists of two main components: the encoder layers and the decoder 
layers. The encoder layers process sequential input data across various time steps in parallel using a multi-head 
self-attention module. They encode input information into contextualized representations. The decoder layers then 
process these encoded representations along with previously produced tokens to generate the corresponding output 
sequences. We refer readers to the literature \cite{transformer, transhealthsurvey} for a detailed explanation of 
the operations performed in each module.

Transformer models exhibit remarkable capabilities in modeling complex linguistic data and text generation. 
Thus, it is extensively used in various NLP tasks \cite{nlp_survey}, including sentiment analysis 
\cite{sentiment}, machine translation \cite{machine_tranlate}, question answering 
\cite{QA}, and information extraction \cite{extract1}. The self-attention mechanism is highly 
parallelizable. This significantly reduces training time and enables large-scale pre-training. As a result, many 
Transformer-based large language models (LLMs), such as OpenAI's GPT series \cite{gpt-3, gpt-4}, Google's 
BERT \cite{bert}, and Meta's LLaMA series \cite{llama, llama-2}, are intensively pre-trained on an enormous corpus of 
text data through self-supervised or semi-supervised learning. These pre-trained LLMs are often referred to as 
foundation models \cite{foundation, foundation2} since they are trained on a broad spectrum of 
unlabeled data, enabling to tackle general tasks like natural language understanding and text generation. These 
foundation models can then be fine-tuned on task-specific data for downstream tasks through transfer learning. 

Transformer models have had a significant impact in other domains as well, such as CV \cite{ViT_survey, ViT4CV, ViT}, remote sensing imaging \cite{remote_sensing}, speech processing \cite{branchformer}, and 
time-series modeling \cite{time-series}.  Furthermore, many LVMs \cite{ViT, clip, convnet} have emerged as 
foundation models in the CV domain. However, the development of foundation models for sensor data domains, especially 
in the field of consumer healthcare, remains largely unexplored. 

\subsection{Parameter-Efficient Fine-Tuning Methods for Foundation Models}
\label{sec:peft}
As foundation models are pre-trained for general-purpose objectives, full fine-tuning is often required to adapt them 
to specific downstream tasks. However, with the increasing size of these models and downstream datasets, fine-tuning 
entire models for every new task has become impractical and resource-intensive. Hence, PEFT methods are crucial for 
adapting foundation models to downstream tasks without consuming extensive computational resources and training time. 
They selectively tune a small subset of parameters, thereby reducing the computational burden and time. Prominent PEFT 
methods include LoRA \cite{lora}, adapter modules \cite{adapter1, adapter2}, and prefix tuning \cite{prefix1, 
prefix2}. LoRA and its variants \cite{dora, cola} approximate weight updates using low-rank decomposition matrices, 
while adapter modules introduce additional layers between existing ones to capture task-specific information. On the 
other hand, prefix tuning prepends learnable vectors to the input sequence to guide model behavior. These methods 
facilitate efficient and scalable adaptation of large models to diverse applications.

Among the PEFT algorithms, LoRA and its variants, Weight-Decomposed Low-Rank Adaptation (DoRA) \cite{dora} and Chain of 
LoRA (CoLA) \cite{cola}, do not alter the model architecture and are easily applicable to models with numerical sensor 
data inputs. In general, for a foundation model with a pre-trained weight matrix $W_0 \in \mathbb{R}^{d \times k}$, 
these methods approximate its update $W_0 + \Delta W$ by representing the variance $\Delta W$ with a low-rank 
decomposition $\Delta W = BA$, where $B \in \mathbb{R}^{d \times r}$, $A \in \mathbb{R}^{r \times k}$, and rank 
$r \ll \text{min}(d, k)$ \cite{lora}. During training, $W_0$ is frozen and does not receive gradient updates, whereas
$B$ and $A$ contain trainable parameters. Therefore, the model's forward pass with an input $x \in \mathbb{R}^{1 \times 
d}$ after fine-tuning yields $x(W_0 + \Delta W) = x(W_0 + BA)$. In addition, DoRA further decomposes $W_0$ into 
$W_0 = \|W_0\|_c \frac{W_0}{\|W_0\|_c}$, where $m = \|W_0\|_c \in \mathbb{R}^{1 \times k}$ is the magnitude vector, 
$\frac{W_0}{\|W_0\|_c} \in \mathbb{R}^{d \times k}$ represents the direction matrix, and $\| \cdot \|_c$ denotes the 
column-wise norm of a matrix \cite{dora}. $m$ is a trainable vector, where a scalar in $m$ defines the magnitude of 
a column vector in $\frac{W_0}{\|W_0\|_c}$. DoRA then trains the direction matrix in the same way as LoRA. Hence, the 
updated forward pass with DoRA yields $x(m'\frac{W_0 + BA}{\|W_0 + BA\|_c})$. On the other hand, CoLA extends LoRA by 
adding a chain of low-rank decomposition matrices to approximate the optimal weight variance $\Delta W^* = 
\sum_{j=1}^l B_jA_j$, where $B_j \in \mathbb{R}^{d \times r_j}$, $A_j \in \mathbb{R}^{r_j \times k}$, and 
$r_j \ll \text{min}(d, k)$ for $j = \{1, 2, \ldots, l\}$ and $l \in \mathbb{N}$. Accordingly, the updated forward 
pass in CoLA yields $x(W_0 + \Delta W^*) = x(W_0 + \sum_{j=1}^l B_jA_j)$.

\subsection{Transformer Models in Smart Healthcare}
Transformer models are gaining increasing attention and adoption in the smart healthcare domain \cite{hllm_survey, 
hllm_5, transhealthsurvey, health1, health2, health3}. Han \textit{et al.} \cite{hllm_3} fine-tune publicly accessible 
pre-trained LLMs on specialized datasets crafted with biomedical domain knowledge. Their fine-tuned model, MedAlpaca, 
consistently outperforms its pre-trained-only counterparts on the United States Medical Licensing Examination (USMLE). 
Singhal \textit{et al.} \cite{hllm_1} demonstrate promising performance on medical question answering datasets with 
their Transformer model, Med-PaLM, developed by adapting a foundation model with instruction prompt tuning. Toma 
\textit{et al.} \cite{hllm_4} introduce Clinical Camel based on LLaMA-2 \cite{llama-2}, an open LLM explicitly 
tailored to clinical research using QLoRA \cite{qlora}. Clinical Camel achieves state-of-the-art (SOTA) performance 
across multiple medical benchmarks relative to openly available medical LLMs and is capable of synthesizing plausible 
clinical notes. Li \textit{et al.} \cite{chatdoctor} adapt and refine LLaMA \cite{llama} using a large dataset of 
100,000 patient-doctor dialogues sourced from a public online medical consultation platform to develop a medical 
chatbot, ChatDoctor. ChatDoctor can understand patient needs and provide informed advice with self-directed 
information retrieval from reliable online sources and offline medical databases. 

In addition to their adaptation in medical NLP applications, Transformer models are reshaping the healthcare
domain by enabling multimodal medical data analysis, e.g., based on medical images and physiological signals. Chen
\textit{et al.} \cite{transunet} propose TransUNet as a strong alternative for medical image segmentation.
TransUNet integrates a U-shaped convolution neural network (CNN) with a vision Transformer. It outperforms
various competing methods on different medical applications, including multi-organ segmentation and cardiac
segmentation. Dai \textit{et al.} \cite{transmed} develop TransMed for multimodal medical image classification.
TransMed combines the advantages of CNN and Transformer to efficiently extract low-level features of images and
establish long-range dependencies between modalities. It achieves superior performance relative to other SOTA
CNN-based models on two medical image datasets. On the other hand, Kim \textit{et al.} \cite{health-llm} extend
the capacity of pre-trained LLMs to deliver multimodal health predictions based on contextual linguistic information 
and physiological data. They fine-tune off-the-shelf LLMs using prompting, instruction tuning, and PEFT on six 
public health health datasets. They demonstrate promising performance on various clinical and wellness prediction 
tasks. However, despite the growing adaptation of Transformers to smart healthcare, the development of a health 
foundation model dedicated to consumer health tasks based on WMS data remains underexplored. 

\section{The COMFORT Framework}
\label{sec:COMFORT}
This section discusses the COMFORT framework in detail. We begin with a top-level overview. Then, we describe the 
health foundation model. Finally, we elaborate on COMFORT's scalability and adaptivity for disease detection 
based on a PEFT method. 

\subsection{Framework Overview}
\begin{figure}[t]
    \centering
    \begin{subfigure}[b]{0.6\textwidth}
        \centering
        \includegraphics[width=\textwidth]{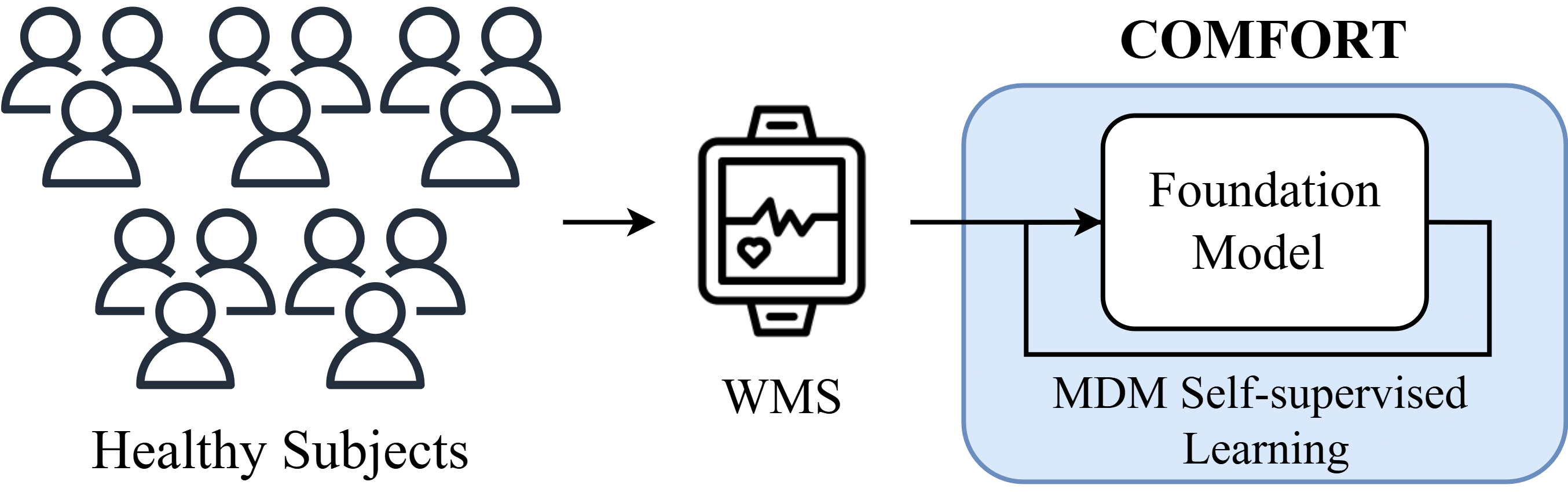}
        \caption{Health foundation model pre-training}
    \end{subfigure}
    \begin{subfigure}[b]{0.7\textwidth}
        \centering
        \includegraphics[width=\textwidth]{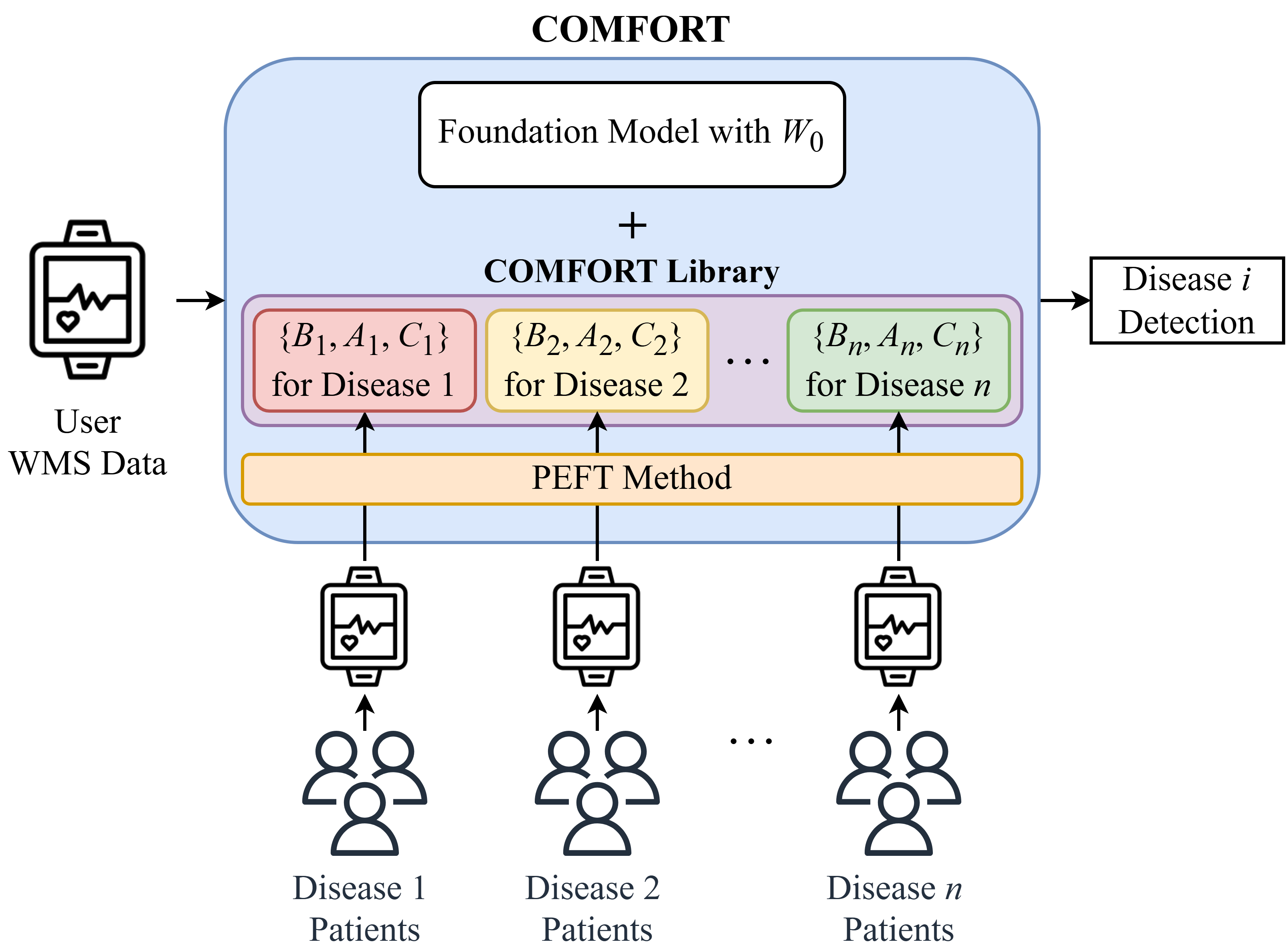}
        \caption{Continual fine-tuning for disease detection}
    \end{subfigure}
    \caption{An overview of the COMFORT framework: (a) health foundation model pre-training, and (b) continual 
fine-tuning for disease detection. $W_0$ represents the foundation model's pre-trained weights.}
    \label{fig:COMFORT}
\end{figure}

Inspired by the success of Transformer-based foundation models in NLP and CV, we propose COMFORT that applies the
pre-train and fine-tune paradigm to the sensor data domain. Fig.~\ref{fig:COMFORT} presents a top-level overview 
of the framework. COMFORT comprises two main components: health foundation model pre-training and continual 
fine-tuning for disease detection. In the pre-training phase, we train a health foundation model on a large dataset 
collected exclusively from healthy individuals using commercially available WMSs. We adopt an MDM self-supervised 
learning objective to enable the foundation model to understand the WMS data and their dynamically-changing 
patterns. During the continual fine-tuning phase, COMFORT keeps the foundation model's pre-trained weights $W_0$ 
frozen. It then applies a LoRA-based \cite{lora} PEFT algorithm to efficiently fine-tune the foundation model for 
downstream detection tasks with disease-specific WMS datasets gathered from patients. Meanwhile, COMFORT maintains 
a library to store all the low-rank decomposition matrices $\{B_i, A_i\}$ and corresponding classifier $\{C_i\}$ 
for all learned diseases. The COMFORT library allows it to continually tackle new disease detection tasks without 
compromising the scalability of the foundation model. Furthermore, it empowers COMFORT's adaptivity to detect 
various diseases by flexibly switching between different $\{B, A, C\}$ triples from the library. Hence, COMFORT 
provides a versatile tool for efficient and effective early-stage disease detection in the consumer healthcare domain. 

\begin{figure}[t]
    \centering
    \includegraphics[width=0.35\textwidth]{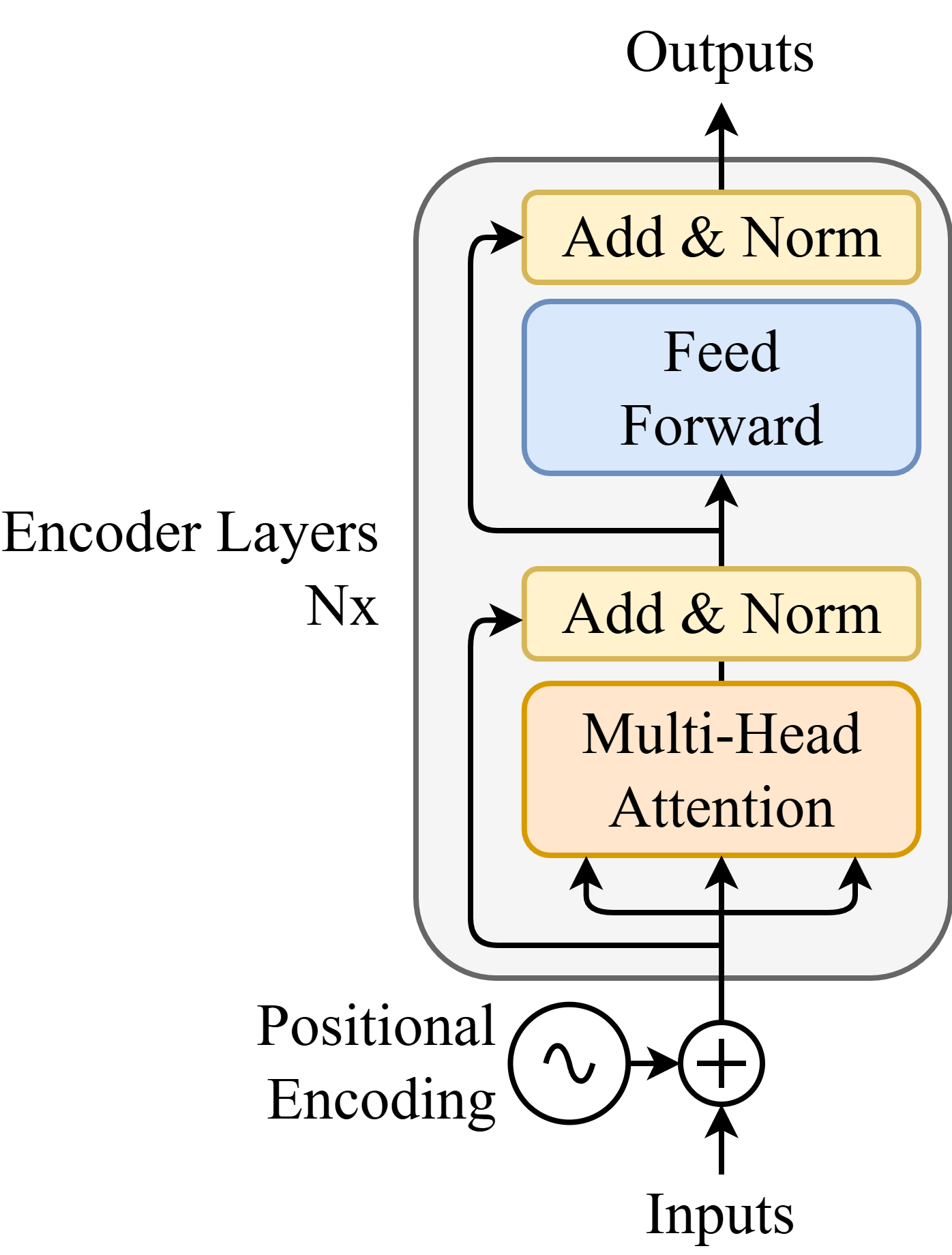}
    \caption{The health foundation model architecture.}
    \label{fig:encoder}
\end{figure}

\subsection{Health Foundation Model}
\label{sec:health}
The self-attention mechanism in Transformer models enables capturing of long-range dependencies within sequential
data and processing of this temporal information in parallel. This makes Transformer models ideal candidates for
analyzing sequential time-series physiological signals collected using WMSs.  The health foundation model in COMFORT
is a Transformer model dedicated to disease detection tasks based on WMS data.  It needs to understand the 
relationship among input sensor data. However, it does not need to generate any generative context except output 
probabilities for inference, which eliminates the need for the decoder layers.  Therefore, we adopt an encoder-only 
Transformer architecture that harnesses bidirectional representations from its self-attention mechanism. We choose 
the BERT \cite{bert} architecture, a multi-layer bidirectional Transformer encoder, for our health foundation model. 
It is almost identical to the original model \cite{bert}. Hence, we follow the method described in \cite{bert} and 
\cite{transformer} to build our model. We omit an exhaustive background description of the model architecture and 
operations. We refer readers to the literature \cite{bert, transformer} for details. Fig.~\ref{fig:encoder} shows 
the model architecture of our health foundation model. Note that we do not need a tokenizer or an input embedding 
module in our model since the input data obtained from WMSs are already in numerical form. 

\begin{figure}[t]
    \centering
    \includegraphics[width=0.8\textwidth]{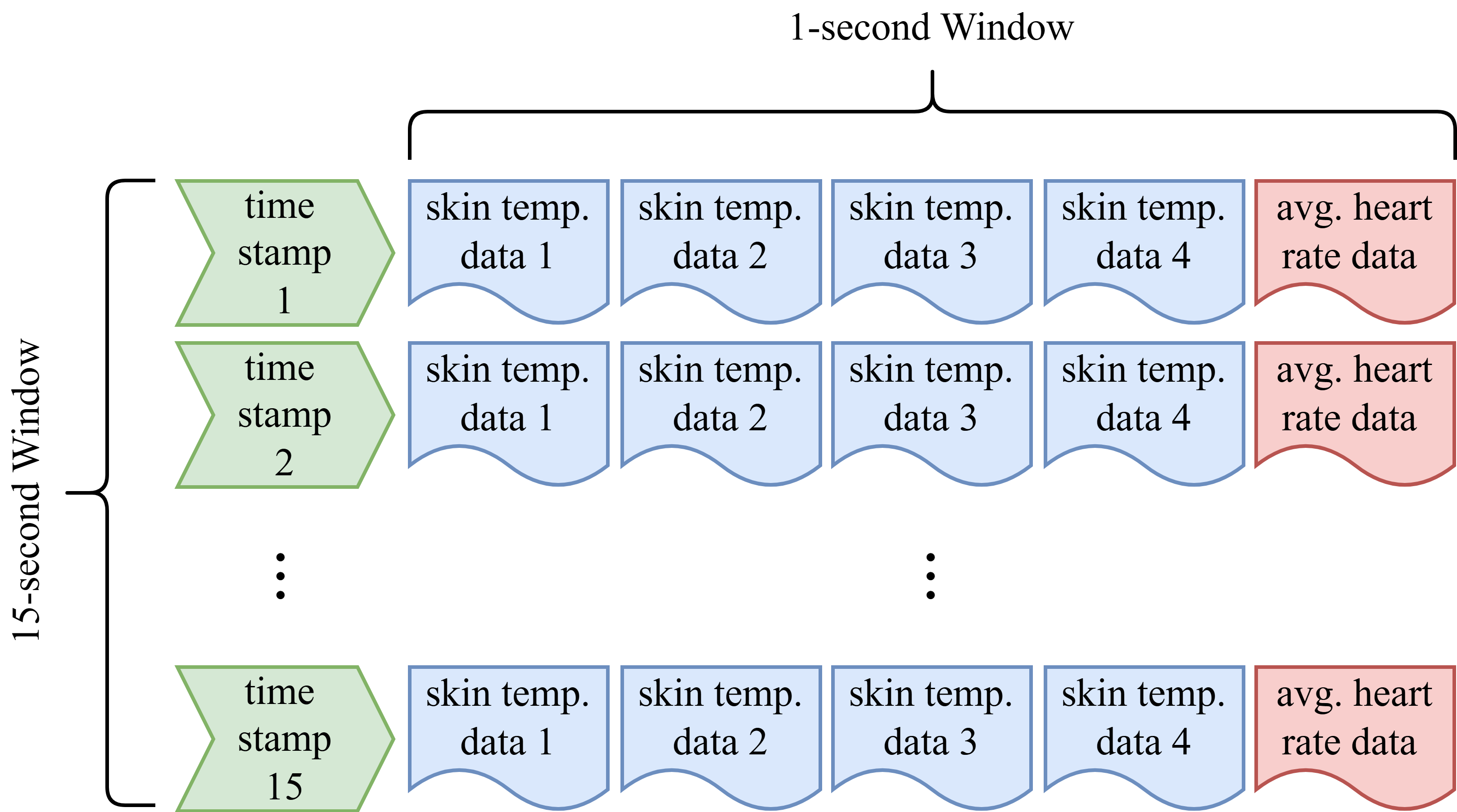}
    \caption{An input data example from a healthy individual. The example data are obtained using a skin temperature 
sensor sampled at 4Hz and a heart rate monitor sampled at 1Hz. (temp.: temperature, avg.: average)}
    \label{fig:input_data}
\end{figure}

Due to data privacy issues and medical regulations, obtaining sufficient patient data to pre-train a foundation model 
for general disease detection tasks is challenging. However, we can potentially collect an unlimited amount of
physiological data from healthy individuals around the world. Therefore, we propose a novel approach of collecting
physiological signals from only healthy individuals to construct a WMS dataset that enables our foundation
model to make sense of the sensor data. We collect essential vital signs, such as galvanic skin response, skin
temperature, average heart rate, and blood volume pulse. These physiological data can be easily collected from
commercially available WMSs like smartwatches, smart rings, or smart wristbands. In addition, we incorporate
motion data and ambient environmental information, including acceleration, orientation, and ambient illuminance,
for healthy individuals using smartphones. These supplementary data have been proven to provide diagnostic
insights through user motion and habit tracking \cite{additional_data, motor_act, temp_health, mhD}. For example,
they can aid in monitoring body movement, detecting physical and mental states, and calibrating physiological
signal collection. We concatenate all WMS and supplementary data from each healthy individual after aligning
them using their time stamps. Next, we segment the data into multiple 15-second windows, where each 15-second
window represents a sequence of 15 consecutive 1-second windows of sensor data. Fig.~\ref{fig:input_data} shows a 
toy example of our input data.  More details of our experimental datasets and data preparation are provided in 
Sections~\ref{sec:datasets} and \ref{sec:preprocessing}.

BERT is pre-trained with a masked language modeling (MLM) objective so that it can learn deep bidirectional representations. 
Similarly, we train our health foundation model to learn bidirectional representations of WMS data by pre-training it on
an MDM self-supervised task. For each epoch, we randomly select five 1-second windows within each 15-second input sequence. 
Within each selected 1-second window, we randomly mask 15\% of the WMS data by replacing them with values randomly sampled 
from a normal distribution. We then use the original, unmasked input sequence as the target for our foundation model to 
enable it to learn to predict the masked values using a mean squared error (MSE) loss criterion.  Through MDM pre-training 
on extensive WMS data collected from healthy individuals, we construct a health foundation model that can serve as a robust 
and adaptable base for various downstream disease detection tasks based on WMS data. 

\subsection{Continual Fine-Tuning for Disease Detection}
COMFORT adapts the health foundation model to downstream disease detection tasks using a PEFT method. As mentioned in 
Section~\ref{sec:peft}, LoRA and its variants, DoRA and CoLA, do not alter the model weights and are easily applicable 
to models with numerical sensor data inputs. Therefore, we select from LoRA, DoRA, and CoLA to implement the PEFT algorithm 
in COMFORT. We compare their performance in our experiments. 

\begin{figure}[t]
    \centering
    \includegraphics[width=0.35\linewidth]{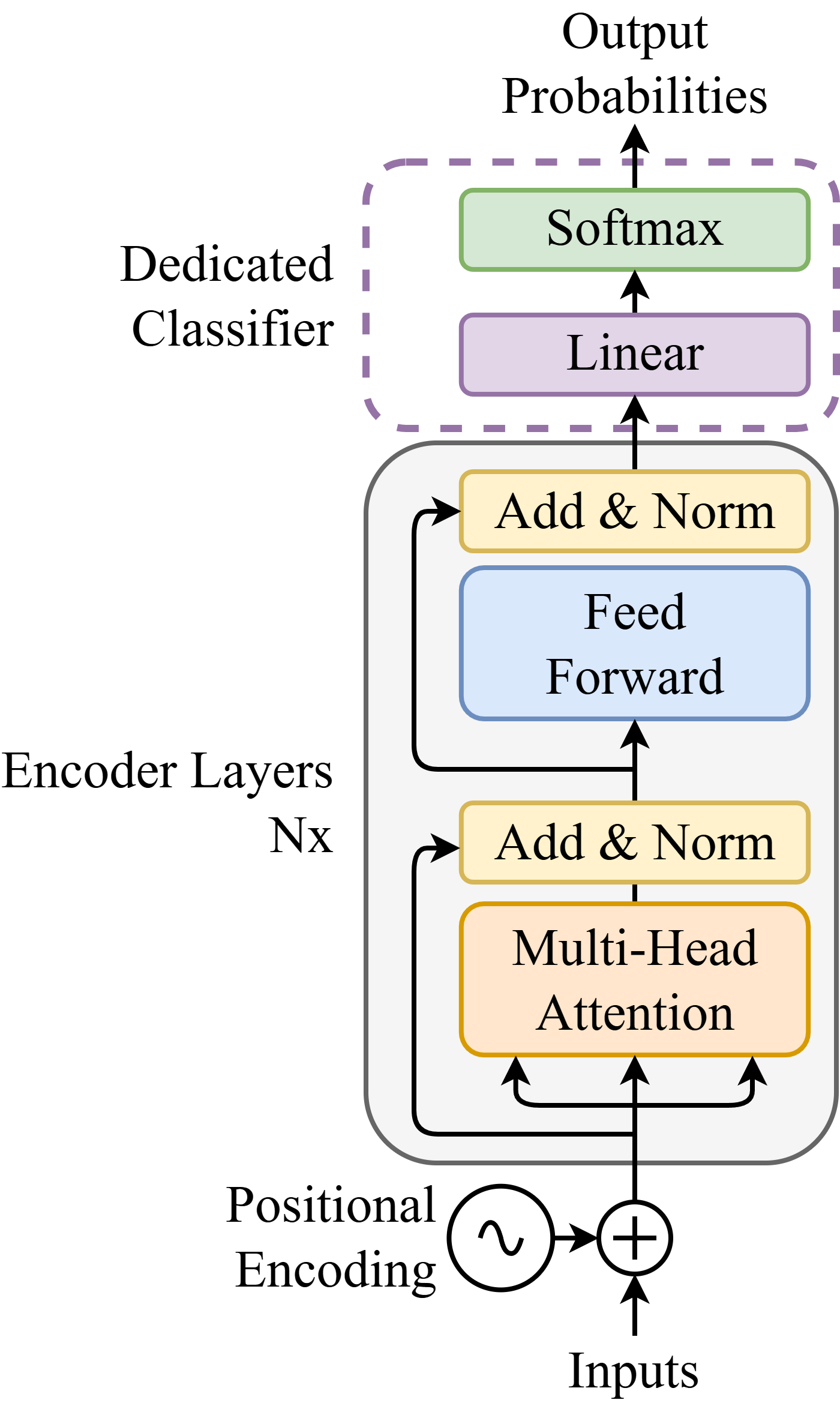}
    \caption{The model architecture after fine-tuning on a downstream task.}
    \label{fig:peft_model}
\end{figure}

We use the original implementation described in the literature \cite{lora, dora, cola} for these PEFT methods. When 
fine-tuning on a new downstream task $i$, COMFORT keeps the health foundation model's pre-trained weights 
$W_0 \in \mathbb{R}^{d \times k}$ frozen. As mentioned in Section~\ref{sec:peft}, the forward passes with an input 
$x \in \mathbb{R}^{1 \times d}$ updated by LoRA, DoRA, and CoLa yield:
\begin{align*}
    \text{LoRA: }x(W_0 + \Delta W_i) &= x(W_0 + B_iA_i); \\
    \text{DoRA: }x(W_0 + \Delta W_i) &= x(m_i'\frac{W_0 + B_iA_i}{\|W_0 + B_iA_i\|_c}); \\
    \text{CoLA: }x(W_0 + \Delta W_i^*) &= x(W_0 + \sum_{j=1}^l B_{ij}A_{ij}),
\end{align*}
\noindent where $\{B_i, B_{ij}\} \in \mathbb{R}^{d \times r}$, $\{A_i, A_{ij}\} \in \mathbb{R}^{r \times k}$,
$m_i' \in \mathbb{R}^{1 \times k}$, and $r \ll \text{min}(d, k)$. We use random Gaussian initialization for
$\{A_i, A_{ij}\}$ and zero for $\{B_i, B_{ij}\}$. Similar to the original implementation, we then scale $\{B_iA_i,
B_{ij}A_{ij}\}$ by $\frac{\alpha}{r}$, where $\alpha$ is a constant. In addition, COMFORT appends a dedicated
classifier $C_i$, comprising a linear module and a softmax layer, to classify disease $i$ during fine-tuning.
Fig.~\ref{fig:peft_model} illustrates the model architecture after fine-tuning for a downstream task. 

In addition, COMFORT maintains a library in order to continually store the low-rank decomposition matrices 
$\{\text{LoRA: }(B_i, A_i) \text{, DoRA: }(m_i', B_i, A_i) \text{, CoLA: }(B_{ij}, A_{ij})\}$ and the dedicated classifier 
$\{C_i\}$ for all learned diseases. Preserving the low-rank matrices instead of the updated weights $\{W_0 + \Delta W_i\}$ 
or the weight variances $\{\Delta W_i\}$ reduces memory overhead significantly. This strategy enables us to 
continually fine-tune the health foundation model for various disease detection tasks. When deployed in production, a user 
can flexibly download and choose among the desired $\{(B_i, A_i) \text{ or } (m_i', B_i, A_i) \text{ or } (B_{ij}, A_{ij}) 
\text{and} C_i\}$ from the COMFORT library to detect disease $i$ on an edge device with commercially available WMSs. 
Therefore, COMFORT provides a scalable and adaptive framework for personalized and proactive consumer health disease 
detection with WMSs. 

\section{Experimental Setup}
\label{sec:setup}
In this section, we provide details of the experimental setup. First, we introduce the datasets used in our experiments. 
Next, we elaborate on the preprocessing procedures applied to these datasets. Finally, we present the implementation 
details of our COMFORT framework.

\subsection{Datasets}
\label{sec:datasets}
The datasets we use in our experiments are from prior studies: DiabDeep \cite{diabD} and MHDeep \cite{mhD}. The data 
collection and experimental procedures for both datasets were approved by the Institutional Review Board of Princeton 
University. The efficacy of both the datasets has been established in the literature \cite{diabD, mhD}. In both datasets, 
we collect distinct physiological signals from participants with a commercially available WMS, an Empatica E4 smartwatch. 
In addition, we collect motion data and ambient environmental information from participants using a Samsung Galaxy S4 
smartphone. These supplementary data have been demonstrated to provide diagnostic insights based on user motion and habit 
tracking in the literature \cite{diabD, mhD}. Table~\ref{tbl:DDFeatures} lists the sensor features collected in these 
datasets and their respective sampling rates, sources, and data types. 

The DiabDeep dataset consists of sensor data collected from 52 participants, including 14 diagnosed with Type-I diabetes, 
13 diagnosed with Type-II diabetes, and 25 non-diabetic individuals. None of the participants report mental, cardiac, or 
endocrine disorders. The MHDeep dataset comprises data collected from 72 adult participants at the Hackensack Meridian 
Health Carrier Clinic in Belle Mead, New Jersey. These participants were diagnosed by medical professionals at the clinic. 
The dataset includes 23 participants with bipolar disorder, 10 with major depressive disorder, 16 with schizoaffective 
disorder, and 23 without any mental health disorder. For both datasets, participants are instructed to wear the Empatica 
E4 smartwatch on the wrist of their non-dominant hand and place the Samsung Galaxy S4 smartphone in the opposite front 
pocket. The data collection process lasts between 1.0 and 1.5 hours per participant, during which time participants are 
allowed to move around freely with the devices.

\begin{table}[t]
    \caption{Sensor Data Collected in the DiabDeep and MHDeep Datasets}
    \centering
    \resizebox{0.90\linewidth}{!}{
    \begin{tabular}{lccc}
    \toprule
    Sensor Feature & Sampling Rate (Hz) & Sensor Source & Data Type \\
    \midrule
    Galvanic Skin Response ($\mu S$) & 4 & \\
    Skin Temperature ($^\circ C$) & 4 & \\
    Acceleration ($x, y, z$) & 32 & Smartwatch & Continuous \\
    Average Heart Rate & 1 & \\
    Blood Volume Pulse & 64 & \\
    \midrule
    Humidity & 5 & \\
    Ambient Illuminance & 5 & \\
    Ambient Light Color Spectrum (R, G, B, W) & 5 & \\
    Ambient Temperature ($^\circ C$) & 5 & \\
    Gravity ($x, y, z$) & 5 & \\
    Angular Velocity ($x, y, z$) & 5 & \\
    Orientation ($x, y, z$) & 5 & Smartphone & Continuous \\
    Acceleration ($x, y, z$) & 5 & \\
    Linear Acceleration ($x, y, z$) & 5 & \\
    Air Pressure & 5 & \\
    Proximity & 5 & \\
    Wi-Fi Radiation Strength & 5 & \\
    Magnetic Field Strength & 5 & \\
    \bottomrule
    \end{tabular}
    }
    \label{tbl:DDFeatures}
\end{table}

\subsection{Dataset Preprocessing}
\label{sec:preprocessing}
We preprocess and format both datasets to fit our foundation model for experiments. First, we synchronize the sensor data 
from the smartwatch and smartphone using their timestamps. Next, we segment the data streams into 15-second windows with a 
15-second shift in between to minimize time correlation between adjacent windows. Each 15-second window of data constitutes 
a data sequence, which we further split into 15 1-second data instances. Analogous to NLP, we format each sensor data 
sequence as a linguistic sequence (sentence) with 15 tokens (words), where each token is encoded as a 1-second data stream 
of all sensor features. Within each 1-second data instance (or token), we flatten and concatenate data from the smartwatch 
and smartphone. This results in a total of 20,957 data sequences in the DiabDeep dataset and 27,082 data sequences in the 
MHDeep dataset, with each data instance containing 299 features. Therefore, the DiabDeep dataset has a [20957, 15, 299]
size, whereas the MHDeep dataset has a [27082, 15, 299] size. 

\begin{figure}[t]
    \centering
    \includegraphics[width=0.90\linewidth]{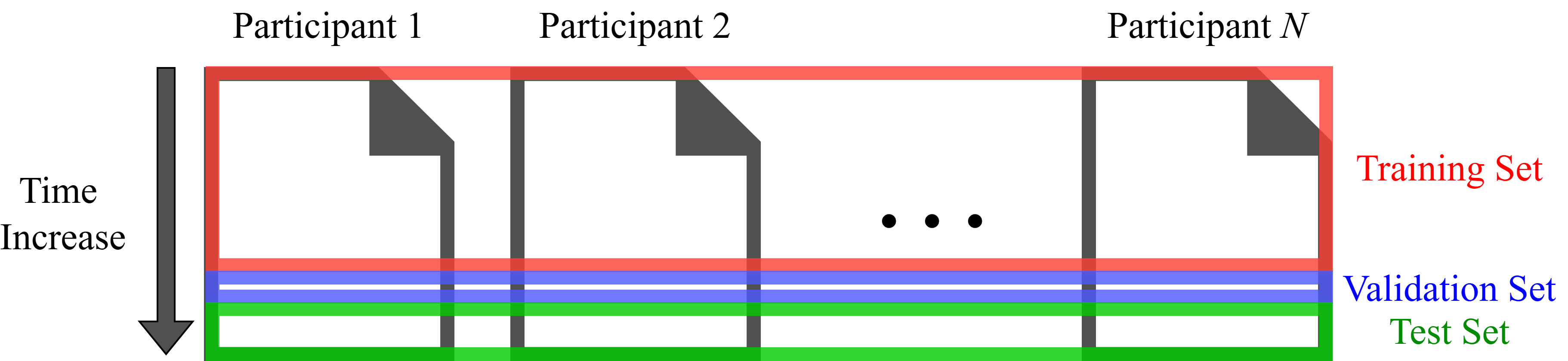}
    \caption{Preparation of the training, validation, and test sets.}
    \label{fig:split}
\end{figure}

Subsequently, we partition each dataset into three subsets for experiments: training, validation, and test sets. We use 
the first 70\%, the next 10\%, and the last 20\% of each participant's sequential time-series data to construct the 
training, validation, and test sets with no time overlap. Fig.~\ref{fig:split} illustrates how we split each dataset for 
experiments. In addition, we perform min-max normalization on both datasets to scale them to the range between 0 and 1. 
This prevents features with a wider range of values from overshadowing those with a narrower range and helps stabilize 
the training process. 

To mitigate the curse of dimensionality and decrease both training time and model size, we use principal component 
analysis (PCA) to reduce the dimensionality of both datasets. To preserve at least 99\% of the total variance while ensuring 
training efficiency, we reduce the dimensionality to 128 based on our empirical results for both datasets. First, we 
standardize the feature data in the training set and compute the covariance matrix. Then, we perform eigendecomposition to 
find the eigenvectors and eigenvalues of the covariance matrix to identify the principal components. Next, we sort the 
eigenvectors in descending order of their corresponding eigenvalues. Finally, we construct a projection matrix to select 
the top 128 principal components and transform the data along these principal component axes. This results in 
128-dimensional feature data in our training set. We then employ the same PCA parameters in the validation and test sets to 
transform them into 128-dimensional feature data as well. 

Finally, we perform data cleaning to remove low-quality data instances from both datasets. Our data are collected
using a WMS worn on the wrist and a smartphone placed in the pocket. Consequently, some data might not be of high
quality when the devices are not properly attached to the wrist of the participant or when the participant is in motion. 
Thus, data cleaning is important for WMS datasets. We use CTRL \cite{CTRL} for data cleaning of both datasets. CTRL was 
originally developed to detect label errors. However, we have discovered through empirical experience that the paradigm 
is also effective in identifying incorrect or corrupted data. CTRL first trains a neural network with the original dataset 
and records the training loss curves of all the samples. Then, K-means clustering is used to the recorded training losses 
to cluster the samples into two groups: clean samples (associated with a smaller training loss centroid) and compromised 
samples (associated with a larger training loss centroid). We use CTRL to clean the training, validation, and test sets 
from both datasets. Table~\ref{tbl:sizes} lists the dataset sizes obtained after applying our preprocessing procedures. 

\begin{table}[t]
    \caption{The DiabDeep and MHDeep Dataset Sizes}
    \centering
    \resizebox{0.85\linewidth}{!}{
    \begin{tabular}{ccccc}
    \toprule
    \multicolumn{2}{c}{\multirow{2}{*}{Dataset}} & \multirow{2}{*}{Original Size} & After Min-Max & After Data Cleaning \\
     & & & Normalization and PCA & with CTRL \\
    \midrule
    \multirow{3}{*}{DiabDeep} & Training Set & [14671, 15, 299] & [14671, 15, 128] & [14438, 15, 128] \\
     & Validation Set & [2098, 15, 299] & [2098, 15, 128] & [2094, 15, 128] \\
     & Test Set & [4188, 15, 299] & [4188, 15, 128] & [3717, 15, 128] \\
    \midrule
    \multirow{3}{*}{MHDeep} & Training Set & [18955, 15, 299] & [18955, 15, 128] & [18267, 15, 128] \\
     & Validation Set & [2707, 15, 299] & [2707, 15, 128] & [2613, 15, 128] \\
     & Test Set & [5420, 15, 299] & [5420, 15, 128] & [5123, 15, 128] \\
    \bottomrule
    \end{tabular}
    }
    \label{tbl:sizes}
\end{table}

\subsection{Implementation Details}
\label{sec:implementation}
COMFORT addresses disease detection on edge devices based on WMS data. We employ the $\text{BERT}_{\text{TINY}}$ 
architecture \cite{bert-tiny,bert-tiny2} for this purpose. We set the number of Transformer encoder layers $L$ to 2, 
the model's hidden size $H$ to 128, and the number of attention heads $A$ to 2. We follow the original implementation in 
\cite{transformer, bert} to set other Transformer parameters. Subsequently, we implement two hidden layers in the linear 
module for each dedicated classifier for downstream tasks. The classifier begins with an input layer with 128 neurons for
alignment with the output embeddings from the encoder layers. Then, it is followed by two hidden layers with 512 and 128 
neurons, respectively. We adopt the rectified linear unit (ReLU) as the nonlinear activation function in the hidden layers. 
Finally, the classifier concludes with an output softmax layer with the number of neurons corresponding to the number of 
classes in the downstream task. Specifically, there are three neurons for the DiabDeep task and four neurons for the MHDeep 
task. This results in less than 529k parameters in the fine-tuned model. 

In COMFORT, we exclusively use WMS data obtained from healthy individuals to pre-train the health foundation model. 
However, we face limitations due to a lack of available large-scale WMS datasets collected from healthy individuals. 
Meanwhile, our experimental datasets do not provide sufficient WMS data from healthy individuals. Therefore, we use a 
synthetic data generation tool, called TUTOR, for this purpose \cite{tutor, doctor, page}. We employ the Gaussian mixture 
model estimation method \cite{doctor, page} to generate 100,000 synthetic data instances from the WMS data of the healthy 
individuals in both datasets. We use these synthetic data to pre-train the health foundation model and to prevent bias 
during fine-tuning. 

For both pre-training and fine-tuning, we use the Adam optimizer with the learning rate initialized at 0.005 and the batch 
size set to 128. We pre-train the foundation model for 1,000 epochs or until convergence. For fine-tuning, we train the 
model with task-specific datasets for 300 epochs. For fine-tuning, we compare LoRA \cite{lora}, DoRA \cite{dora}, and 
CoLA \cite{cola}. We set both the rank number $r$ and the constant $\alpha$ to eight for all three PEFT methods. In addition, 
we set the chain length to three with a fixed rank number for CoLA. We implement COMFORT with PyTorch and perform 
experiments on an NVIDIA A100 GPU. To accelerate the experiments, we employ the CUDA and cuDNN libraries. 

\section{Experimental Results}
\label{sec:experiments}
This section presents experimental results on fine-tuning for two downstream disease detection tasks: DiabDeep and MHDeep. 
Before fine-tuning, we pre-train the health foundation model on the synthetic healthy individual data, as described in 
Section~\ref{sec:implementation}. We pre-train the foundation model with the MDM objective described in 
Section~\ref{sec:health} until we obtain a loss value lower than 0.001. 

\subsection{Continual Fine-Tuning for Disease Detection}
\label{sec:results}
We compare the performance of LoRA, DoRA, and CoLA with a vanilla full fine-tuning method, where the COMFORT library stores 
the fully updated weight variances $\Delta Ws$. In addition, we implement two different baselines for comparison. Following 
the original implementation \cite{diabD, mhD}, we recreate the DiabDeep and MHDeep multilayer perceptron (MLP) models with 
synthetic data pre-training and grow-and-prune neural network synthesis. It is important to note that the synthetic data 
used for training the DiabDeep and MHDeep MLP models are generated from \textit{all} the data in the training set. On the 
other hand, we build two task-specific BERT\textsubscript{TINY} models that are trained from scratch with Xavier 
initialization \cite{Xavier} for the two tasks, respectively. To evaluate COMFORT's performance, we report the test 
accuracy and F1-score on the learned tasks. The F1-score is computed by defining true positives (negatives) as the 
unhealthy (healthy) data sequences correctly classified as disease-positive (healthy) and false positives (negatives) as 
the healthy (unhealthy) data sequences misclassified as disease-positive (healthy). In addition, we report the model memory 
consumption required to detect the two learned tasks. 

Table~\ref{tbl:result} shows the experimental results after continually fine-tuning for the two downstream tasks, DiabDeep 
and MHDeep. Table~\ref{tbl:breakdown} provides a breakdown of the model memory consumption required. To detect both 
diseases, the system needs to store either both DiabDeep and MHDeep MLP models, two task-specific BERT\textsubscript{TINY} 
models, or the COMFORT framework, which includes the health foundation model and COMFORT library. As we can see from 
Table~\ref{tbl:result}, all Transformer-based models outperform the original MLP models in both test accuracy and F1-score 
for both tasks. This demonstrates the advantage of Transformer models in processing sequential time-series WMS data over 
conventional MLP models, even though MLP models consume less memory. Among the Transformer models, COMFORT achieves 
competitive performance with PEFT methods compared to training task-specific models from scratch and full foundation model 
fine-tuning. Moreover, COMFORT saves more than 34\% and 52\% in memory consumption with LoRA relative to using two 
task-specific models and full fine-tuning, respectively. 

\begin{figure}[t]
    \centering
    \includegraphics[width=0.8\linewidth]{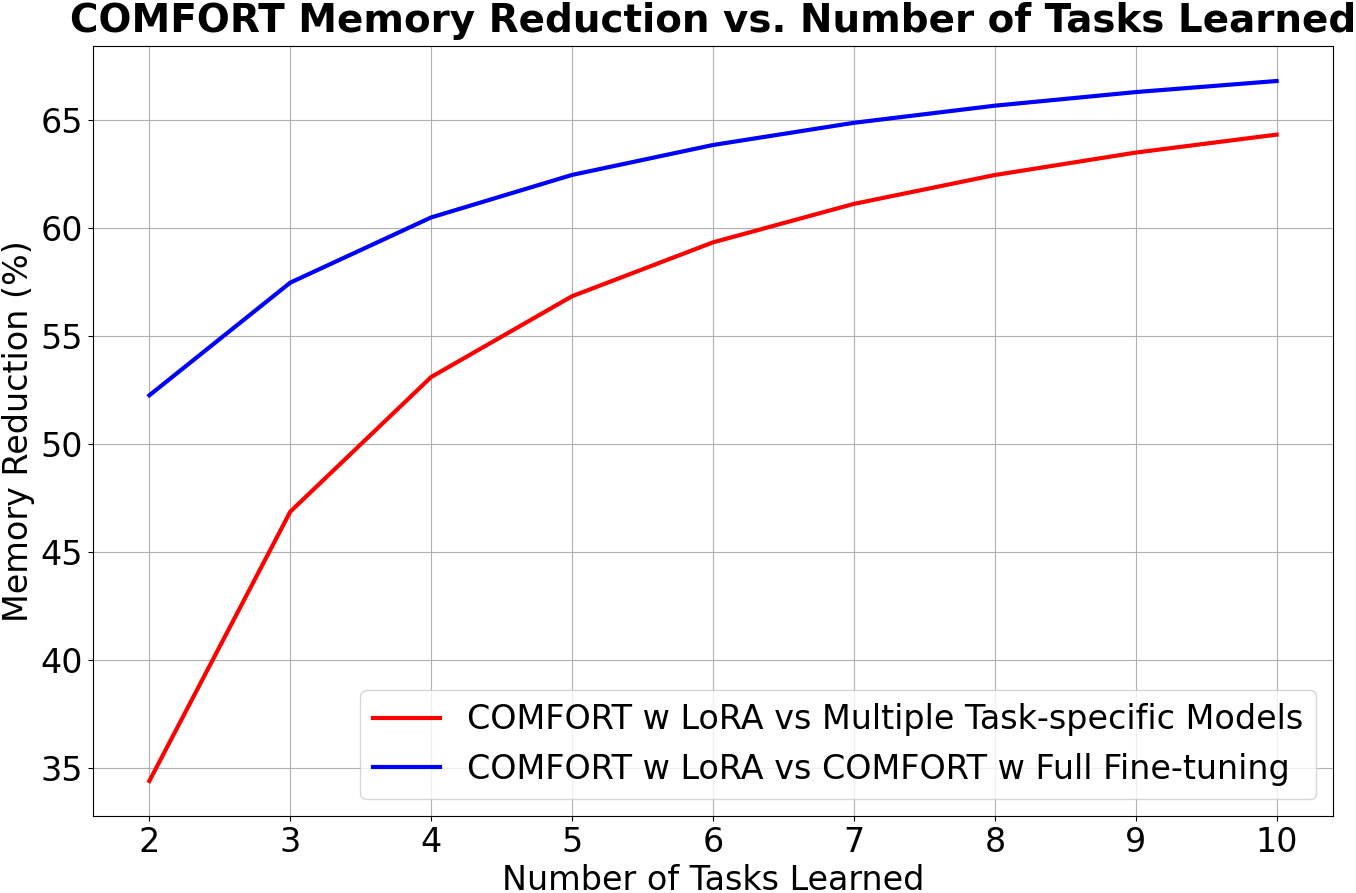}
    \caption{Projected memory reduction when using COMFORT with LoRA after learning 10 downstream tasks.}
    \label{fig:mem_reduc}
\end{figure}

Our experimental results demonstrate that COMFORT creates a scalable and adaptive framework for consumer health
disease detection tasks using WMS data. The advantage of reduced memory consumption becomes more pronounced as more downstream
tasks are learned. 
We simulate a scenario where the health foundation model is continually adapted to 10 distinct downstream disease detection tasks, each involving three-way classification. Fig.~\ref{fig:mem_reduc} shows the projected memory reduction achieved by adopting COMFORT with LoRA for these 10 tasks. The results show that COMFORT paves the way for personalized and proactive healthcare solutions to efficient and effective early-stage disease detection.

\begin{table*}[t]
    \caption{Continual Fine-Tuning for Disease Detection}
    \centering
    \resizebox{1\linewidth}{!}{
    \begin{tabular}{cccccccc}
    \toprule 
    \multirow{3}{*}{Downstream} & \multirow{3}{*}{Evaluation} & Original & BERT\textsubscript{TINY} & \multicolumn{4}{c}{COMFORT} \\
    \cmidrule{5-8}
    \multirow{2}{*}{Task} & \multirow{2}{*}{Metrics} & MLP & Train from & Full & PEFT with & PEFT with & PEFT with \\
     & & Model & Scratch & Fine-tuning & LoRA & DoRA & CoLA \\
    \midrule
    \multirow{2}{*}{DiabDeep} & Test Accuracy & 0.952 & 0.978 & 0.977 & 0.978 & 0.977 & 0.976 \\
     & F1-score & 0.952 & 0.979 & 0.979 & 0.980 & 0.979 & 0.976 \\
    \midrule
    \multirow{2}{*}{MHDeep} & Test Accuracy & 0.892 & 0.939 & 0.938 & 0.927 & 0.932 & 0.927 \\
     & F1-score & 0.978 & 0.998 & 0.995 & 0.992 & 0.991 & 0.994 \\
    \midrule
    \midrule
    (To Detect & Model Memory & \multirow{3}{*}{1,331} & \multirow{3}{*}{4,278} & \multirow{3}{*}{5,878} & \multirow{3}{*}{2,806} & \multirow{3}{*}{2,816} & \multirow{3}{*}{2,946} \\
    DiabDeep \& & Consumption & \\
    MHDeep) & (KB) & \\
    \bottomrule
    \end{tabular}
    }
    \label{tbl:result}
\end{table*}

\begin{table*}[t]
    \caption{Breakdown of Model Memory Consumption}
    \centering
    \resizebox{1\linewidth}{!}{
    \begin{tabular}{ccccccc}
    \toprule 
    (To Detect & Original & BERT\textsubscript{TINY} & \multicolumn{4}{c}{COMFORT} \\
    \cmidrule{4-7}
    $\text{DiabDeep}_{i=1}$ \& & MLP & Train from & Full & PEFT with & PEFT with & PEFT with \\
    $\text{MHDeep}_{i=2})$ & Model & Scratch & Fine-tuning & LoRA & DoRA & CoLA \\
    \midrule
     & $\text{MLP}_1(665)$ & $\text{BERT}_{\text{TINY}\_1}$ & $W_0(1,600)$ & $W_0(1,600)$ & $W_0(1,600)$ & $W_0(1,600)$ \\
    Breakdown & \multirow{3}{*}{$+$} & (2,139) & $+$ & $+$ & $+$ & $+$ \\
    of Memory & & $+$ & $\Delta W_1(2,139)$ & $\{B_1, A_1, C_1\}(603)$ & $\{m'_1, B_1, A_1, C_1\}(608)$ & $\{B_{1j}, A_{1j}, C_1\}(673)$ \\
    Consumption & & $\text{BERT}_{\text{TINY}\_2}$ & $+$ & $+$ & $+$ & $+$ \\
    (KB) & $\text{MLP}_2(666)$ & (2,139) & $\Delta W_2(2,139)$ & $\{B_2, A_2, C_2\}(603)$ & $\{m'_2, B_2, A_2, C_2\}(608)$ & $\{B_{2j}, A_{2j}, C_2\}(673)$ \\
     & & & & & & (for $j = 1,2,3$) \\
    \bottomrule
    \end{tabular}
    }
    \label{tbl:breakdown}
\end{table*}

\subsection{Ablation Study}
We conduct two ablation studies on the COMFORT framework. First, we assess whether the pre-trained health foundation model 
within COMFORT reduces the amount of task-specific training data required for fine-tuning. Next, we evaluate the impact of 
using different rank values in the PEFT algorithms on test accuracy. 

\begin{figure}[t]
    \centering
    \begin{subfigure}[b]{0.8\textwidth}
        \centering
        \includegraphics[width=\textwidth]{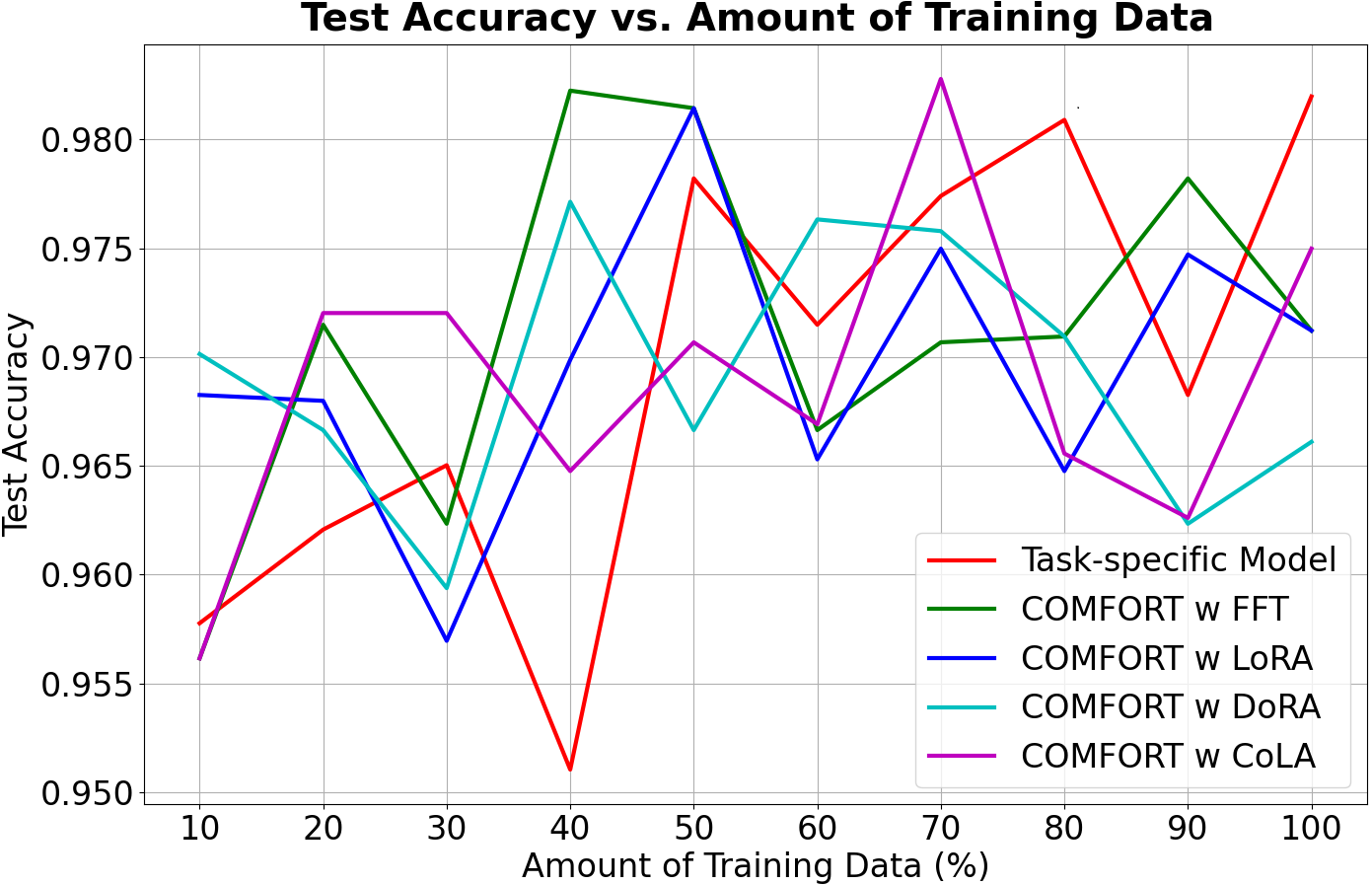}
        \caption{The DiabDeep Task}
    \end{subfigure}
    \begin{subfigure}[b]{0.8\textwidth}
        \centering
        \includegraphics[width=\textwidth]{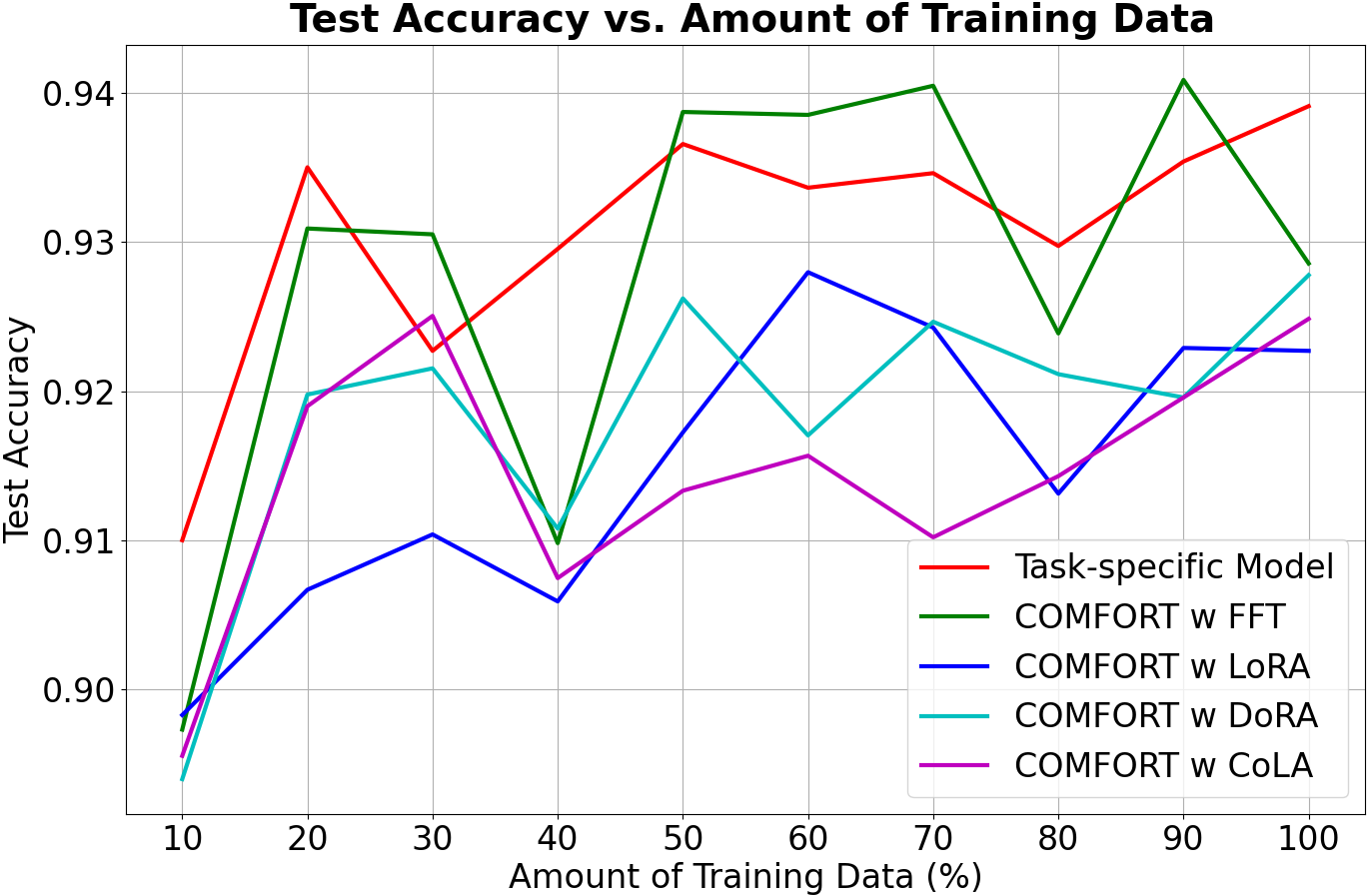}
        \caption{The MHDeep Task}
    \end{subfigure}
    \caption{Test accuracy vs. the amount of task-specific training data used for fine-tuning for the (a) DiabDeep and (b) MHDeep tasks. (FFT: Full fine-tuning.) (Best viewed in color.)}
    \label{fig:abl1}
\end{figure}

Fig.~\ref{fig:abl1} presents the results of the first ablation study: test accuracy versus the amount of task-specific 
training data used for fine-tuning. In this ablation study, we only compare the Transformer models. The $x$-axis indicates 
the percentage of available task-specific training data, while the $y$-axis represents the test accuracy. The results 
show that COMFORT achieves its highest test accuracy with less task-specific training data, regardless of the fine-tuning 
method. For the DiabDeep (MHDeep) task, COMFORT only requires 40\% (70\%), 50\% (60\%), 40\% (50\%), and 70\% (30\%) of the 
task-specific training data for fine-tuning when using the full fine-tuning, LoRA, DoRA, and CoLA methods, respectively. 
In contrast, the task-specific model trained from scratch requires 100\% of the task-specific training data to reach its 
maximum test accuracy on both tasks. This ablation study demonstrates that the health foundation model in COMFORT helps 
reduce the amount of task-specific training data needed for fine-tuning on downstream tasks.

\begin{figure}[t]
    \centering
    \begin{subfigure}[b]{0.8\textwidth}
        \centering
        \includegraphics[width=\textwidth]{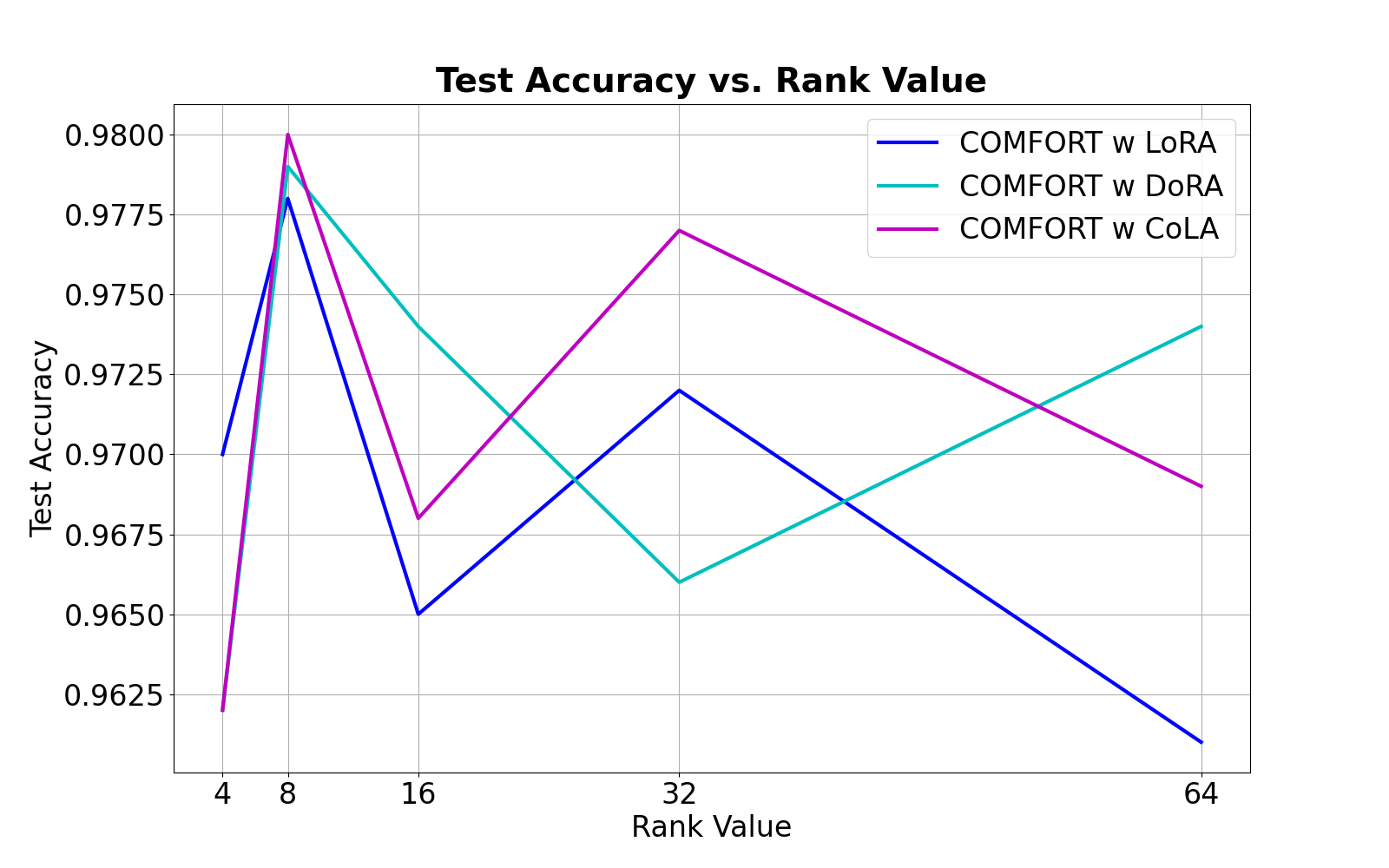}
        \caption{The DiabDeep Task}
    \end{subfigure}
    \begin{subfigure}[b]{0.8\textwidth}
        \centering
        \includegraphics[width=\textwidth]{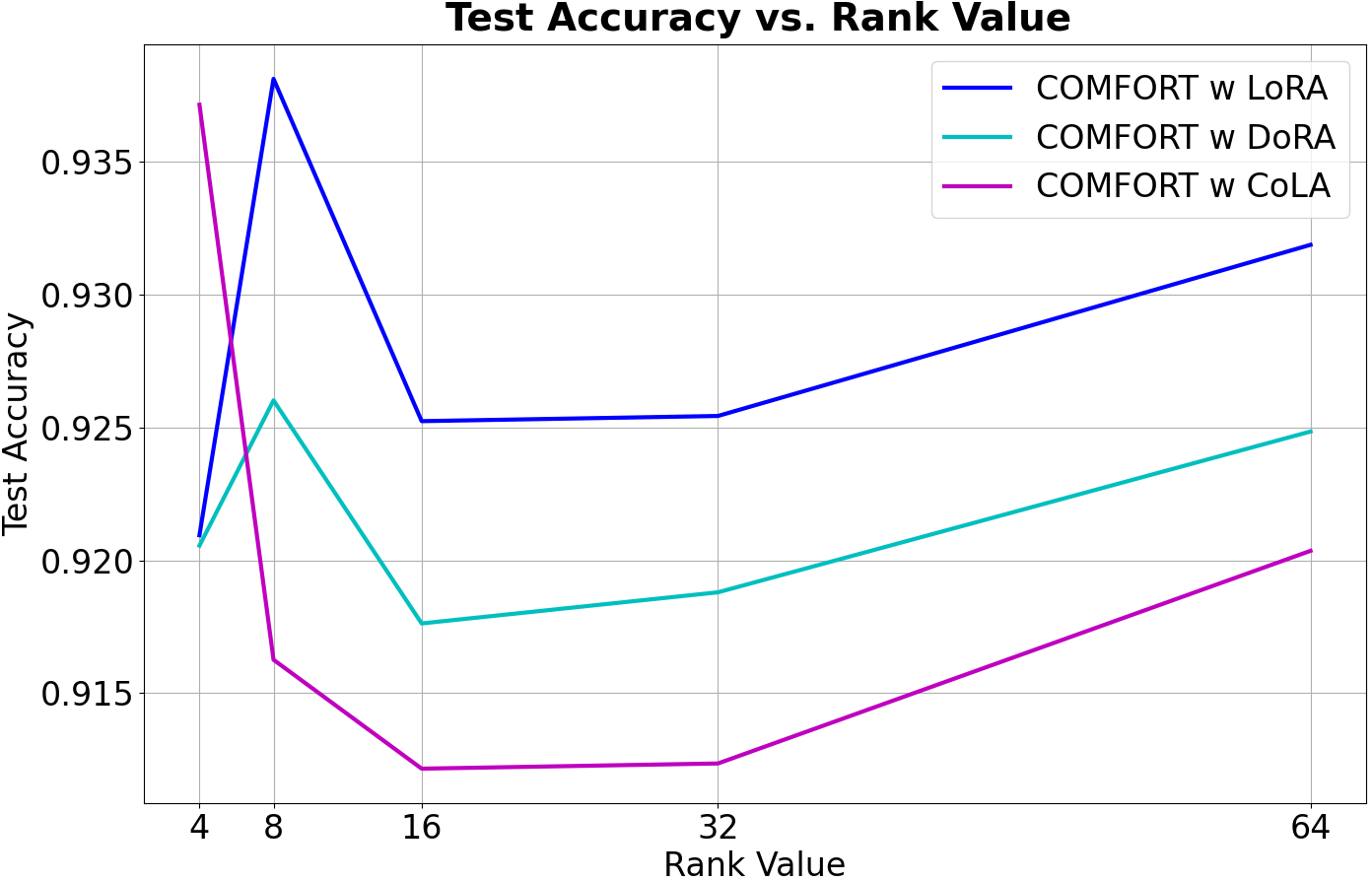}
        \caption{The MHDeep Task}
    \end{subfigure}
    \caption{Test accuracy vs. different rank values for the PEFT method. (Best viewed in color.)}
    \label{fig:abl2}
\end{figure}

Fig.~\ref{fig:abl2} shows the results of the second ablation study: test accuracy versus different rank values used in the 
PEFT algorithms. The $x$-axis denotes the rank values for the PEFT algorithms, whereas the $y$-axis represents the test 
accuracy. For the DiabDeep task, all three PEFT algorithms achieve their highest test accuracy at a rank value of 8. For 
the MHDeep task, both LoRA and DoRA achieve their highest test accuracy when their rank values are set to 8, while CoLA 
reaches its highest test accuracy with a rank value of 4.

\section{Discussions}
\label{sec:discuss}
The premise behind COMFORT is that collecting WMS data from healthy individuals is more feasible than from diagnosed 
patients. Moreover, we can potentially collect unlimited data from healthy individuals worldwide. Therefore, we propose 
foundation models for consumer healthcare applications that undergo extensive pre-training on data collected from healthy 
individuals. This approach mitigates the challenge associated with lack of sufficient patient data needed to train 
foundation models for general disease detection tasks. Our experimental results validate this premise. They
demonstrate the promising potential of extending this approach to other medical data modalities, including medical text and 
images. The COMFORT framework can then be adapted to other smart healthcare applications, such as medical question 
answering, AI chatbots, image segmentation, image classification, and disease detection with medical images. We will 
explore these directions in our future work. 

\section{Conclusion}
\label{sec:conclusion}
In this paper, we described COMFORT, a continual fine-tuning framework for foundation models in the consumer healthcare
domain. It provides a scalable and adaptive solution to consumer health disease detection tasks using WMS data. We
proposed a novel approach to pre-training a health foundation model with data collected exclusively from healthy 
individuals. Subsequently, the foundation model adapts to downstream tasks efficiently using a PEFT algorithm. By storing 
only the low-rank matrices, COMFORT employs a scalable library for continual fine-tuning and flexible disease detection. 
Our experimental results demonstrate COMFORT's efficacy while maintaining a low memory overhead relative to the conventional 
method that employs multiple task-specific models. Hence, COMFORT offers a promising solution to efficient and effective 
early-stage disease detection in the consumer healthcare domain. 

\begin{acks}
This work was supported by NSF under Grant No. CCF-2203399. We would like to thank Sayeri Lala for developing code to 
prepare some of the datasets (specifically mental health and diabetes) used in this study and for her assistance in 
clarifying details concerning the data format.
\end{acks}

\bibliographystyle{ACM-Reference-Format}
\bibliography{biblio}

\end{document}